\documentclass[]{spie}  

\usepackage{amsmath,amsfonts,amssymb}
\usepackage{graphicx}
\usepackage{multirow,tabularx, booktabs}
\newcolumntype{Y}{>{\centering\arraybackslash}X}
\usepackage[colorlinks=true, allcolors=blue]{hyperref}
\usepackage{appendix}

\title{Progressive GANomaly: Anomaly detection with progressively growing GANs}

\author[a]{Djennifer K. Madzia-Madzou}
\author[a]{Hugo J. Kuijf}
\affil[a]{Image Sciences Institute, UMC Utrecht, Heidelberglaan 100, Utrecht, the Netherlands}

\authorinfo{Further author information: (Send correspondence to D.K.M.)\\D.K.M.: E-mail: d.madzia-madzou@students.uu.nl\\  H.J.K.: E-mail: h.kuijf@umcutrecht.nl, Telephone: +31 (0)88 7558562}

\begin{document} 
\maketitle

\begin{abstract}
In medical imaging, obtaining large amounts of labeled data is often a hurdle, because annotations and pathologies are scarce. Anomaly detection is a method that is capable of detecting unseen abnormal data while only being trained on normal (unannotated) data. Several algorithms based on generative adversarial networks (GANs) exist to perform this task, yet certain limitations are in place because of the instability of GANs. This paper proposes a new method by combining an existing method, GANomaly, with progressively growing GANs. The latter is known to be more stable, considering its ability to generate high resolution images. The method is tested using Fashion MNIST, Medical Out-of-Distribution Analysis Challenge (MOOD), and in-house brain MRI; using patches of size 16x16 and 32x32. Progressive GANomaly outperforms a one-class SVM or regular GANomaly on Fashion MNIST. Artificial anomalies are created in MOOD images with varying intensities and diameters. Progressive GANomaly detected the most anomalies with varying intensity and size. Additionally, the intermittent reconstructions are proven to be better from progressive GANomaly. On the in-house brain MRI dataset, regular GANomaly outperformed the other methods.  
\end{abstract}


\section{Introduction}

Since hospitals increase their use of imaging equipment\cite{e_stats}, radiologists experience significantly higher rates of burnout\cite{Harolds2016}. In the past 15 years, the individual workload of radiologists has quadrupled\cite{Bruls2020}, which causes more diagnostic errors\cite{Hames2019}. This ranges on average from 3\% to 5\%\cite{Itri2018}. Out of these errors, 70\% are classified as perceptual errors, where the abnormality is missed. The remaining 30\% are cognitive errors, where the abnormality is detected yet an incorrect diagnosis is given\cite{Berlin2014}. Observer fatigue is a common phenomenon amongst radiologists, since a large volume of images need to be screened with low probabilities of finding a true positive\cite{Thrall2018}. The use of artificial intelligence and machine learning, for example by pre-analyzing the medical images, has the potential to lessen this increasing workload\cite{Thrall2018}. 

Recently, the introduction of generative adversarial networks (GANs)\cite{Goodfellow2014} gained some traction in medical image analysis. Applications include: denoising\cite{Wolterink2017denoising}, reconstruction\cite{Zhang2018}, conditional synthesis\cite{Wolterink2017synthesis}, registration\cite{Yan2018}, unconditional synthesis\cite{Baur2018syn}, segmentation\cite{Son2017}, classification \cite{Ren2018}, and detection\cite{Baumgartner2018}. Using GANs, a general detection method through anomaly detection can be created. Instead of training a method to look for a specific disease, an anomaly detector has the potential to identify new patterns, which are not present in the training data\cite{Pimentel2014, Markou2003}. 

Anomaly detection in medical image analysis is a very young cross-sectional field\cite{Schlegl2017}. Examples of other domains where this method is applied include credit-card fraud detection \cite{Awoyemi2017} and X-ray security imaging of luggage\cite{Akcay2020}. These algorithms learn the distribution of training data, which are labeled as ‘normal’, and build a representation of these distributions in latent space. A new image is labeled as ‘abnormal’, if the position of the image is far from the known representations in said space. One of the first methods to perform this task on (medical) images is the one-class support vector machine (SVM)\cite{Scholkopf1999, Singh2012, Seebock2016, Quellec2016, Giritharan2009, Bowles2017}. More recent methods are based on (variational) autoencoders and GANs. Autoencoders have been applied on brain MR images for the detection of lesions\cite{Baur2018seg, Chen2018}. GANs have been applied on optical coherence tomography images\cite{Schlegl2017, Schlegl2019}, on brain MRI for the detection of tumor tissue\cite{Sun2020, Alex2017}, white matter hyperintensities (WMH)\cite{Bowles2017, Kuijf2016}, acute and chronic brain infarcts\cite{Alex2017, Hespen2021}, and multiple sclerosis lesions\cite{Baur2020steGANomaly}. GANomaly\cite{Akcay2015} is one of these architectures, which for certain tasks outperforms other methods such as AnoGAN\cite{Schlegl2017} and f-AnoGAN\cite{Schlegl2019}.

\subsection{Technical Background}
\subsubsection{Generative Adversarial Network} A generative adversarial network is an unsupervised architecture that is based in game theory with two opposing parties: when one agent gains another loses\cite{Goodfellow2014}. A GAN consists out of a generator and a discriminator. The generator creates images from random vectors, whereas the discriminator has access to the training set that contains real images. The discriminator will try to separate the generated images from the real and thus will be optimized to recognize the fake images. As the discriminator is improving, the generator has to produce more realistic fake images. This iterative cycle makes sure that both networks improve. The ultimate goal of a GAN is to produce realistic looking generated images. 

One of the most popular GAN network designs is the deep convolutional generative adversarial network (DCGAN)\cite{Radford2016}. The DCGAN has its own strength and limitations, for instance it is able to produce sharp images, but only for fairly small resolutions. Additionally, the DCGAN is known to be unstable to train owing to the nature of these networks. Common failures observed are vanishing gradients, mode collapse, and failure to converge. To improve the stability, the Wasserstein loss and gradient penalty have been introduced\cite{Arjovsky2017, Gulrajani2017}.

\subsubsection{Progressively growing GANs} 
\begin{figure}[tb]\centering
	\includegraphics[width=0.8\textwidth]{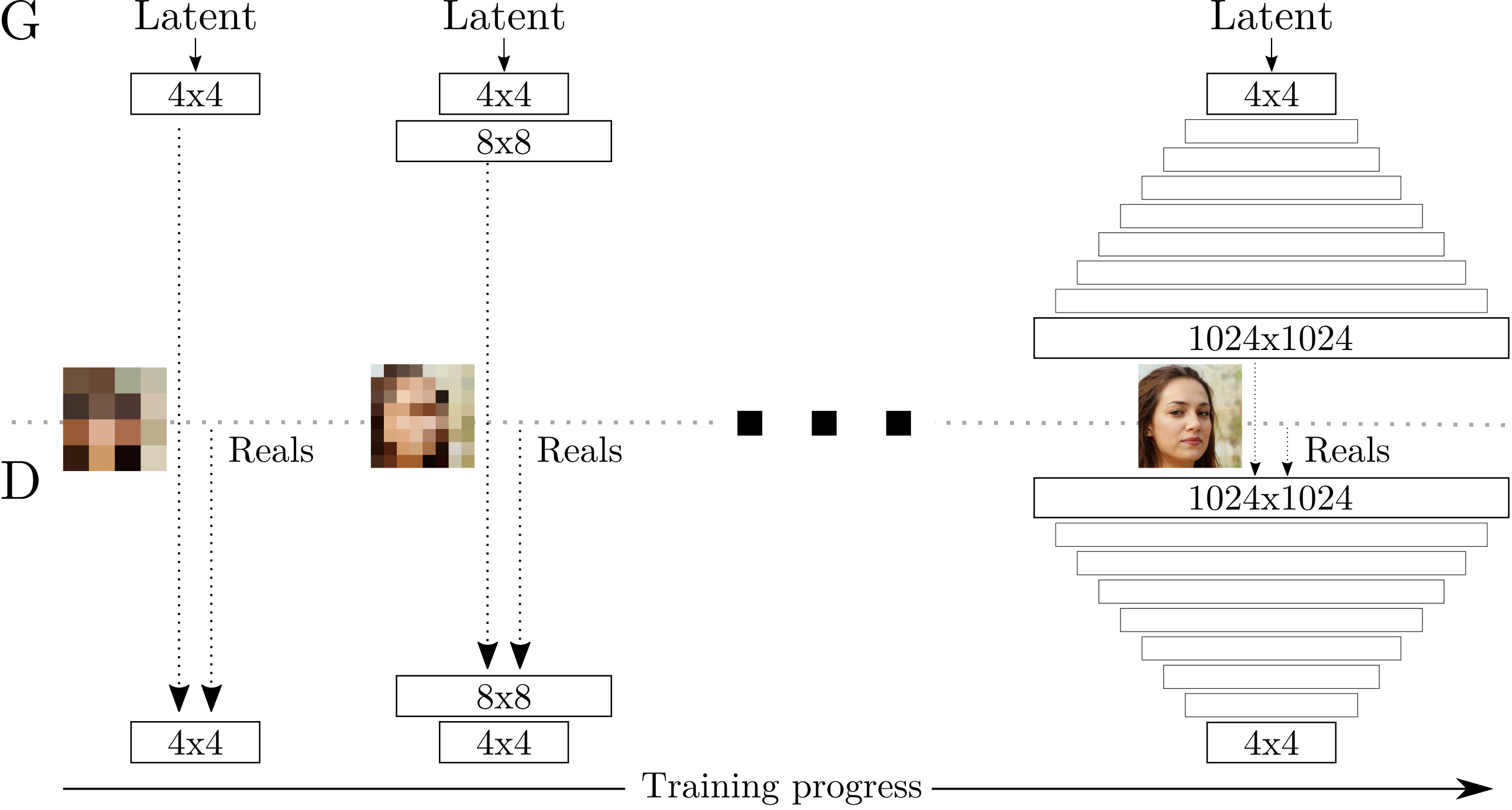}
	\caption{Training for progressively growing GANs, where the process first starts with the Generator (G) and Discriminator (D) training on low resolution images. As training progresses, a layer is added to both G and D to accompany the increasing resolution. Adapted from Progressive growing of GANs for improved quality, stability, and variation by Karras, T., Aila, T., Laine, S., \& Lehtinen, J., 2017, ArXiv, 1–26.}
	\label{fig:pgan}
\end{figure}
As the spatial resolution of images increases, the difficulty to successfully train a GAN increases accordingly. This is because its easier for the discriminator to tell the generated images apart from the real images in high spatial resolutions\cite{Odena2017}. To make training a GAN for high resolution images possible, Karras et al. proposed a progressive growing of GANs\cite{Karras2017}. The major change in this algorithm is that the training starts with a very low resolution. When both the generator and discriminator have been trained sufficiently, the resolution of the images double in each dimension. With this improvement, very realistic images of celebrities have been generated at a resolution of 1024x1024 (Figure \ref{fig:pgan}). 

This work proposes an improvement of GANomaly by replacing the WGAN by a progressively growing GAN. This method will henceforward be called progressive GANomaly. The algorithm per layer in progressive GANomaly is built upon Wasserstein GAN with gradient penalty (WGAN-GP). The best result in the lower resolution needs to be saved and blended with the currently training layer of the higher resolution. This happens by fading the new layers in smoothly with the older ones. The proposed methods will be subjected to a series of experiments incrementing in complexity. First both GANomaly and progressive GANomaly will be tested on Fashion MNIST against a baseline method, one-class SVM. Subsequently, the GAN based methods will be tested and compared using two different brain datasets, MOOD and an in-house dataset. 

\section{Methods}
\subsection{One-class SVM}
A baseline method is used to compare the potential of GANomaly and progressive GANomaly as anomaly detectors. One-class SVM is an unsupervised outlier detection algorithm. It learns a decision function to classify new data as similar or different to the training data. This algorithm uses a hypersphere to encompass all training data with the largest possible margin\cite{Tax2001}. If new data falls outside of the hypersphere, it will be classified as abnormal. A radial basis function (RBF) kernel is used, which defines the hypersphere in a non-linear fashion.

\subsection{GANomaly}
In order to turn a GAN into an anomaly detector, changes are needed in the architecture. The discriminator normally outputs a single scalar value, signifying the realness of an image. In GANomaly, the discriminator has three different outputs: 1) A single scalar output to predict if the image is real or fake. 2) The discriminator is used as an encoder to produce a latent vector from an image. This latent vector represents the image in a lower dimensionality. 3) The features of an image are encoded in the intermediate layer of the discriminator. Using these features to train the generator has been shown to reduce the instability of the GAN training\cite{Salimans2016}. The generator does not alter and decodes an image from a latent vector.

\begin{figure}[tb]\centering
	\includegraphics[width=0.8\textwidth]{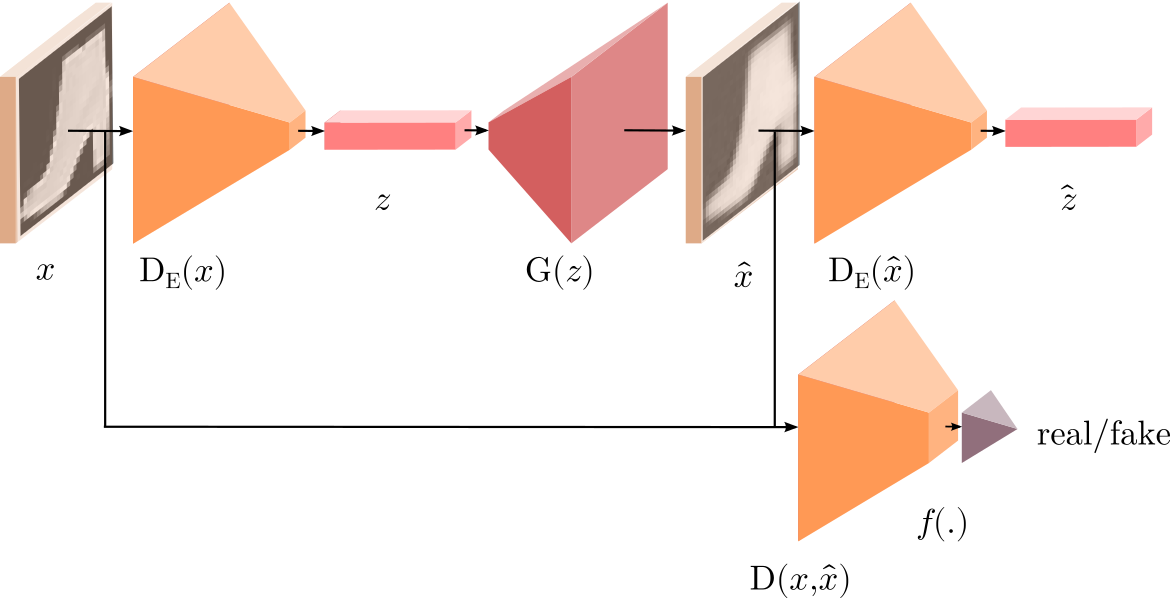}
	\caption{Pipeline of GANomaly using an image and its reconstruction from Fashion MNIST as an example.}
	\label{fig:pipeline}
\end{figure}

The pipeline of GANomaly is shown in Figure \ref{fig:pipeline} and starts with input image \(x\). This image is encoded by the discriminator \(D_E(x)\) into latent vector \(z\). The generator \(G(z)\) translates latent vector \(z\) into a reconstructed image \(\hat{x}\). This reconstructed image \(\hat{x}\) gets encoded \(D_E(\hat{x})\) into another latent vector \(\hat{z}\) by the same discriminator. Subsequently, the features \(f(.)\) of both input image \(x\) and the reconstruction \(\hat{x}\) are extracted by the intermittent layers of the discriminator \(D(x,\hat{x})\) before a real/fake classification is given.

To train the generator of GANomaly, three loss functions are used. The first one is the encoder loss:
\begin{equation}
	L_{enc} =\lVert z-\hat{z} \rVert_{2}
	\label{eq:enc}
\end{equation}

Following the contextual loss:
\begin{equation}
	L_{con} =\lVert x-\hat{x} \rVert_{1}
	\label{eq:con}
\end{equation}

Lastly the adversarial loss:
\begin{equation}
	L_{adv} =\lVert f(x)-f(\hat{x}) \rVert_{2}
	\label{eq:adv}
\end{equation}

To train the discriminator, the following loss functions are minimized: the binary cross entropy of the predicted label of a real image and the expected value, and the binary cross entropy of the predicted label of a reconstructed image and the correct value. These errors are averaged and used to improve the discriminator. The performance of the GAN is improved by using Wasserstein loss\cite{Arjovsky2017}.

\subsubsection{Parameters}
The network uses minibatches of size 128. The latent vector size is 100. The learning rate is set to 0.002, with Adam as the optimizer\cite{Kingma2015}. The weights associated with the adversarial, contextual, and encoder losses are 1, 70 and 10, respectively. The network keeps training iteratively until convergence in the train-validation set for an established amount of epochs. The epoch with the lowest loss is used to save the weights of the model.

\subsubsection{Anomaly scoring}
The anomaly scores are computed after the model has been trained. A test image is fed into the pipeline and the encoder loss is calculated for that image. This encoder loss is the anomaly score of the test image. During training, the training images are looped through the network to calculate the median and median absolute deviation per element of the encoder loss. These values are used to calculate a modified anomaly score\cite{Hespen2021}, by subtracting the median and dividing by the median absolute deviation.

\begin{figure*}[tb]\centering
	\includegraphics[width=0.8\textwidth]{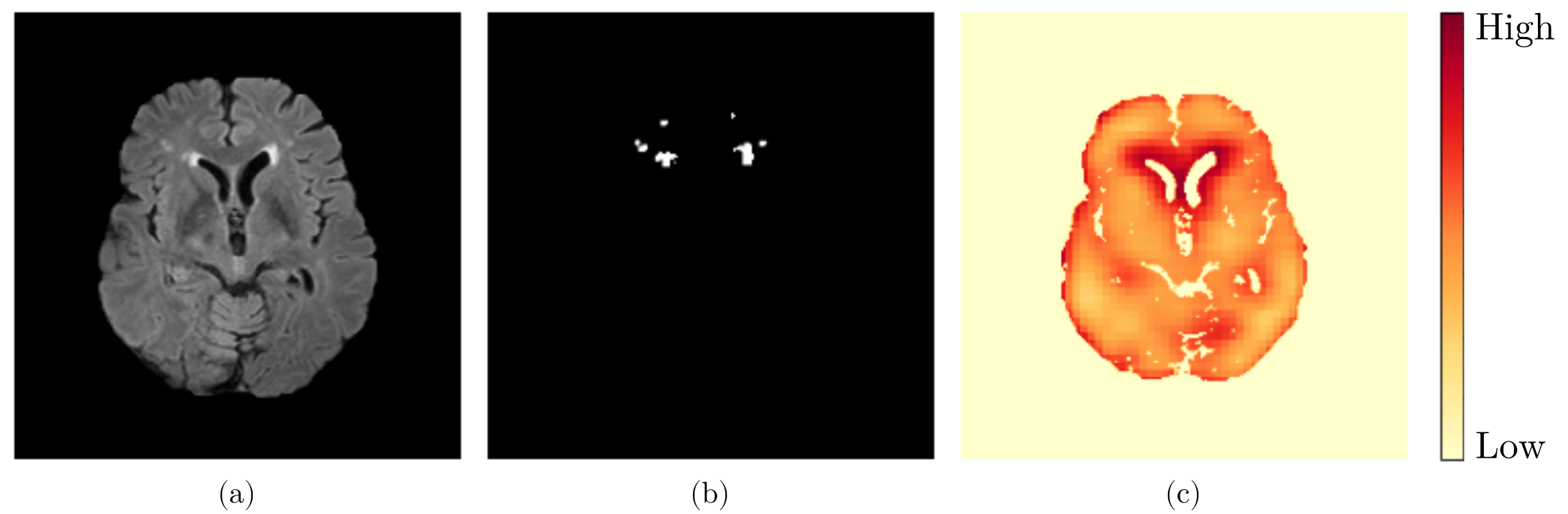}
	\caption{(a) An example of the in-house brain MRI dataset after brain extraction and intensity normalization. The FLAIR images shows white matter hyperintensities (WMH) as bright spots. (b) Manual annotation of WMHs. (c) Using progressive GANomaly with patch size 16x16, a heat map is computed that corresponds to the resulting anomaly score. A red color indicated a high score and the yellow tones a low score.}
	\label{fig:UDES}
\end{figure*}

In case of larger images, a patch based approach will be used, since the method works best in lower resolutions. The image will be deconstructed into overlapping transversal patches with stride 4. The patches will go through the pipeline. Each patch will be reconstructed, assigned an anomaly score and normalized. This results in an anomaly score given on a pixel level of a larger image (example in Figure \ref{fig:UDES}).

\subsection{Progressive GANomaly}
Progressive GANomaly retains the structure of GANomaly, but instead of using a Wasserstein GAN a progressive growing GAN. This is composed out of Wasserstein GAN with gradient penalty\cite{Gulrajani2017}. This algorithm will be trained starting from a resolution of 8x8 and ending with the chosen maximum resolution (16x16 or 32x32). The parameters are kept the same to GANomaly to provide a somewhat objective comparison. Progressively growing GANs do have an additional parameter \(\alpha\) that controls the blending of the different layers. At the start of a new resolution \(\alpha\) grows linearly from 0 to 1 during the first 750,000 iterations. After the initial iterations, the layers are fully blended and continue training with \(\alpha\) equal to 1. The anomaly score is computed by taking the encoder loss without the modifications mentioned in GANomaly. The stopping criterium, that makes GANomaly stop training, is used in progressive GANomaly to jump to a higher resolution. Here, the last best epoch is reloaded and training with an added layer resumes. If the stopping criterium is reached at the wanted maximum resolution, the training will be terminated and the best epoch saved. 

\section{Data and experiments}

\subsection{Fashion MNIST}
Within the computer vision domain, standardized datasets are used to develop various image processing and analysis systems. One of these standardized datasets is Fashion MNIST\cite{xiao2017/online}. This dataset is used for benchmarking algorithms. Fashion MNIST consists out of a selection of Zalando’s articles. The training set contains 60,000 examples, where each image is a 28x28 grayscale image with one of ten labels. Each labels stands for an article of clothing. Machine learning classification techniques easily achieve 97\% accuracy on the regular MNIST\cite{xiao2017/online}, which is why Fashion MNIST was chosen over MNIST. Before training, these images are bilinearly interpolated to size 32x32. This is done because progressively growing GANs only take image sizes which are powers of two.

\subsubsection{Experiment}
\begin{figure}[tb]\centering
	\includegraphics[width=0.8\textwidth]{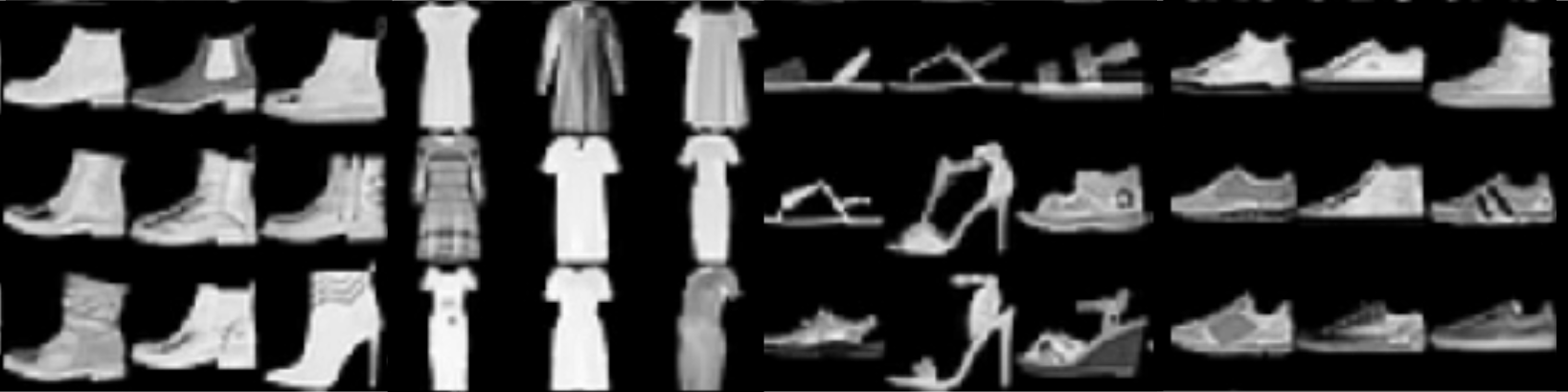}
	\caption{Examples of the Fashion MNIST dataset. The labels that are shown from left to right are: boots, dresses, sandals, and sneakers.}
	\label{fig:fashionmnist}
\end{figure}

To test the anomaly detector algorithm, the model will be trained on one label of Fashion MNIST. This label will be considered as ``normal''. In this case, the boots are normal. After building a model that has learned the distribution of boots, several tests will be performed on different items of clothing. The anomaly scores of boots are going to be compared to dresses first. These two items are far apart in latent space and are therefor expected to be easier to distinguish\cite{fashionmnistweb}. The same will be done with boots and sandals, and boots and sneakers. The latter is expected to be the most challenging one, since these images visually look the most similar. Figure \ref{fig:fashionmnist} offers visualizations of all the used labels. One-class SVM will be used as a baseline method and it will be compared to both GANomaly and progressive GANomaly.

For the one-class SVM, 80\% of boots are used for training. The remaining 1200 boots were used for test, together with 1200 randomly dresses, sandals, or sneakers. For the methods based on GANs, the training data is split in four ways. There are 6,000 boots available in the dataset. 70\% of these images are used as training data and 5\% are used as a train-validation during training to prevent overfitting. The test-validation group consists of 5\% of the images. This group is used to decide the threshold for the anomaly scores. The last 20\% are used for test, together with 1200 items from one of the other labels. So the final test set is a 50/50 split between anomalous (not boots) and normal (boots) images.

The metrics that will be used to evaluate the performance of these image level detection methods are accuracy, sensitivity, precision, and F1-score. 

\subsection{MOOD}
The second dataset used is from the Medical Out-of-Distribution Analysis Challenge (MOOD)\cite{mood_2021_4573948}, which contains 800 brain MRI images of size 256x256x256. 600 images are used during training. Per image, 2,000 random brain patches are made of size 16x16 and 32x32. 80\% of those patches are used for training and 20\% for train-validation. During training a batch size of 4,200 training patches and 300 validation patches is used. These patches are randomly sampled from the total amount of patches before every epoch. This dataset is trained until the methods have shown to converge in the last 6 epochs.

\subsubsection{Experiment}
After training the GANs to be able to reconstruct normal brain patches, which will be evaluated using the structural similarity index measure (SSIM), the algorithms will be tested with various toy images. One 3D MRI image is taken and a disk with a diameter of 35 mm is inserted into the axial plane in 50 slices with different intensities per image (see Figure \ref{fig:mood}). The intensity of the disk linearly increases between 0.1 and 1.0. For reference, white matter in these images has an intensity between 0.4 and 0.5, grey matter has an intensity of roughly 0.2 and 0.3, and the fat surrounding the brain has an intensity in the region of 0.7. These toy images will be tested using GANomaly and progressive GANomaly with a control image. Another 3D MRI image is used to make toy images with varying disk diameters from 1 to 10 mm in 50 slices using intensity 1.0. Both tests will be evaluated with Receiver Operating Characteristics (ROC) curves to measure the degree of separability of the methods using different thresholds for the anomaly scores. During this evaluation, the background will not be taken into account in the true positive rate. So the ROC curve and Area Under Curve (AUC) score will be reflecting the performance within the brain specifically. 

\subsection{In-house brain MRI}
The last dataset that is analyzed includes 120 in-house brain MR images with manual annotations of white matter hyperintensities (WMH). Scans were acquired using a 3~T MRI system (Philips Medical Systems, Best, the Netherlands) with a standardized protocol including, among others, a FLAIR sequence and a 3D T1-weighted sequence. These two images are used to detect WMHs using anomaly detection. The brain is extracted using the HD-BET tool\cite{Isensee2019}. Subsequently, the images are normalized so that the 5th percentile is set to zero and the 95th percentile to one. Figure \ref{fig:UDES} shows a slice of the FLAIR brain image after preprocessing and the corresponding WMH delineations. 90,000 training patches are extracted of 90 subjects with the least WMHs. While creating the patches, the WMHs are not incorporated to make sure to build a training set of normal patches of both size 16x16 and 32x32 voxels. Out of the 90,000 patches, 80\% is used as training set and 20\% is used as train-validation. The training set is augmented by randomly flipping the patches horizontally. The training stops when the loss of the train-validation set does not improve for ten epochs. For GANomaly this means that it stops training and progressive GANomaly will first jump to a higher resolution before it stops at the desired image size, which is either 16x16 or 32x32.

\subsubsection{Experiment}
Both algorithms are evaluated using the remaining 30 subjects with the most WMHs. These images are preprocessed identically to the training set. Subsequently, they are deconstructed into patches with stride 4. An anomaly score is given to each patch and reconstructed into the full image. The anomaly scores will be evaluated with ROC curves and the corresponding AUC scores, excluding the background. 

\section{Results}

\subsection{Fashion MNIST}
The comprehensive evaluation of all the methods is provided in Appendix Table \ref{tab:fashion}. Testing the one-class SVM, GANomaly, and progressive GANomaly on boots and dresses, while trained on boots, results in F1-scores of 79.8 for the SVM, 99.5 for GANomaly, and 99.6 for progressive GANomaly. When testing on sandals, the scores are lower with F1-scores of 79.7, 94.8, and 95.8, for SVM, GANomaly, and progressive GANomaly, respectively. Comparing boots and sneakers resulted in the lowest scores. SVM got an F1-score of 77.7, GANomaly 81.2, and progressive GANomaly 83.2. In all cases progressive GANomaly reached the highest F1-scores.

\subsection{MOOD}

After training, a visual difference in the quality of the reconstructions is observed between GANomaly and progressive GANomaly. This is confirmed with the SSIM between normal test patches and their corresponding reconstructions. Test brain patches that are reconstructed using GANomaly and 16x16 patches have an SSIM of 0.64. When using progressive GANomaly, this is increased to 0.83. The same increase is observed when using 32x32 patches: GANomaly obtains an SSIM of 0.46 while progressive GANomaly reaches 0.81.

\subsubsection{Intensity test}
\begin{figure}[tb]\centering
	\includegraphics[width=0.4\textwidth]{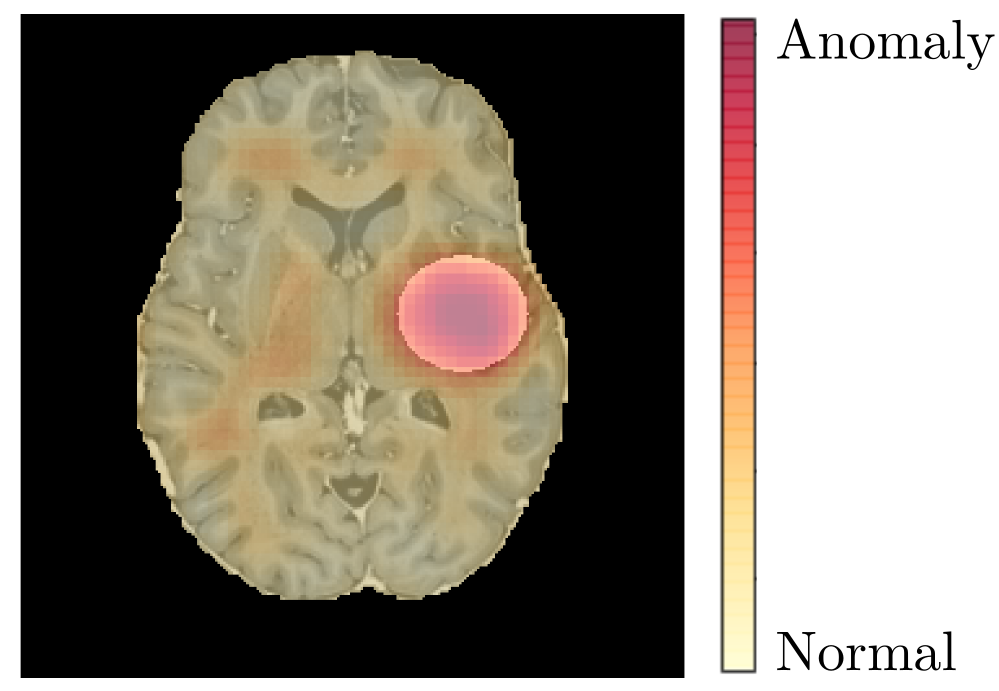}
	\caption{An example of the result computed by progressive GANomaly with patch size 16x16 of a toy image used for the intensity test. A disk with intensity 1 and diameter 35 mm is inserted in the brain image. This image shows the original input image with a heat map representing the given anomaly scores.}
	\label{fig:mood}
\end{figure}

Appendix Table \ref{tab:mood} shows the exact AUC scores per method, patch size, and intensity. Bright disks (intensity over 0.8) are easy to detect for all the methods, yielding a perfect score. Comparing the 16x16 patches of GANomaly and progressive GANomaly, it can be observed that both detect the same amount of intensities, yet different ones. GANomaly additionally detects disks with intensity 0.7 and progressive GANomaly disks with intensity 0.5 (which corresponds to white matter). In the control images, white matter received a slightly higher score compared to the rest of the brain, which can be observed in Figure \ref{fig:mood}. With this intensity test it is also seen that progressive GANomaly is able to detect this specific intensity perfectly. Using larger patches of 32x32, the methods overall detect abnormalities with more different intensities. 32x32 GANomaly detects disks with intensity of 0.6 or higher; and progressive GANomaly with intensity of 0.5 or higher. It is also observed that the larger patches better enable detection of darker anomalies. For GANomaly, this result is seen with intensity 0.1 and progressive GANomaly with intensity 0.2. The anomalies with other intensities are not detectable using these algorithms.

\begin{figure*}[tb]\centering
	\includegraphics[width=0.95\textwidth]{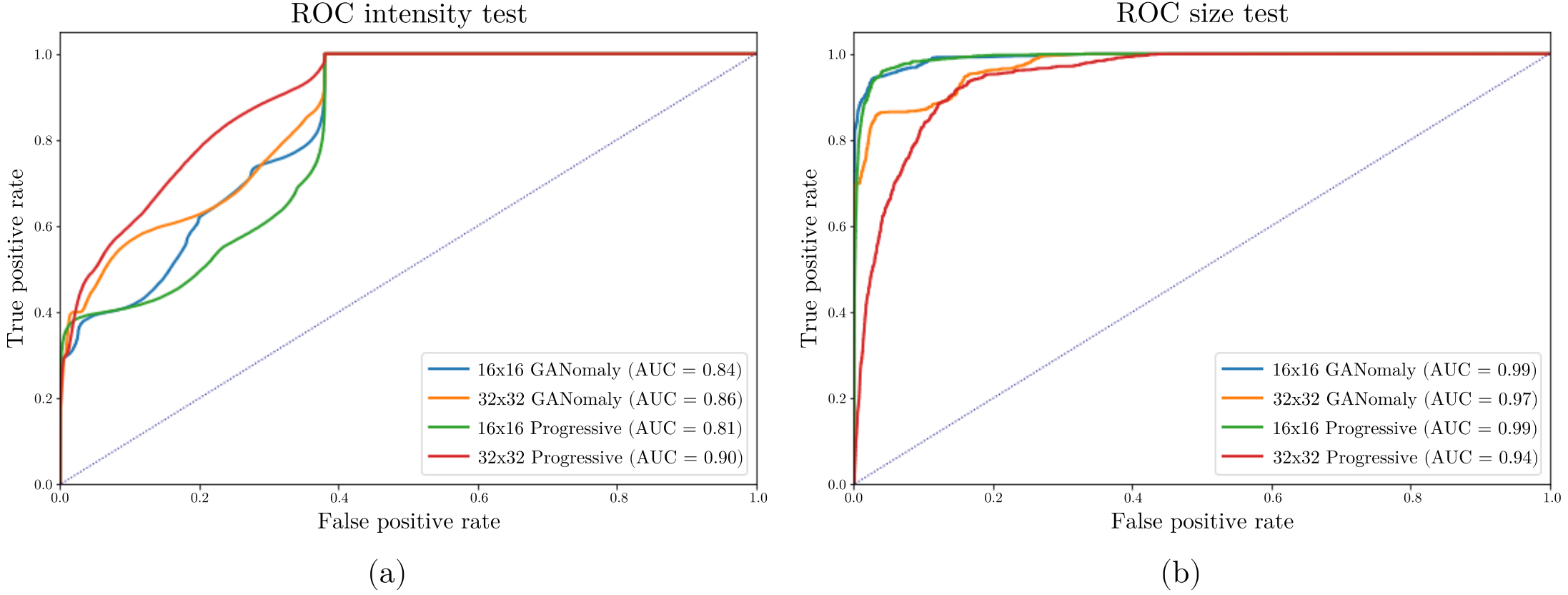}
	\caption{Visual summary of the tests performed on MOOD. On the left side (a) the overall ROC scores are visible per method of the intensity test over all the test images. On the right (b) the same plot is visible for the size test. Here, only disk sizes of 1 mm to 7 mm are taken into account since the bigger anomalies are classified perfectly by every method. On the x-axis the true positive rate is given and the y-axis shows the false positive rate. The legend shows which color corresponds to which method and patch size. It also presents the AUC score over all the images per method. The diagonal blue line shows a model with an AUC of 0.5, which means that it has no discrimination capacity to distinguish between the given classes.}
	\label{fig:roc}
\end{figure*}

No trends were found in the AUC scores of these methods. The overall performance on all intensities of these methods and patch sizes is visualized in Figure \ref{fig:roc}.

\subsubsection{Size test}
Appendix Table \ref{tab:mood} lists all the AUC scores from the tested methods and patch sizes at varying disk sizes. Disks with a diameter larger than 7~mm are perfectly detected by all methods. Using larger patch sizes makes the methods less sensitive to small anomalies. GANomaly using 32x32 patches does receive a higher AUC score for all the different sizes, compared to progressive GANomaly with the same patch size. When using the 16x16 patches, the performance of progressive GANomaly is similar to GANomaly. Progressive GANomaly does seem to be better able to detect the smallest anomalies in this test. The overall performance of both methods during the size test is shown in Figure \ref{fig:roc}. In the Figure, only disk sizes between 1 and 7~mm are taken into account, to better visualize the differences between the methods (thus ignoring the perfect results for disks larger than 7~mm). 

\subsection{In-house brain MRI}
\begin{figure}[tb]
\centering
	\includegraphics[width=0.49\textwidth]{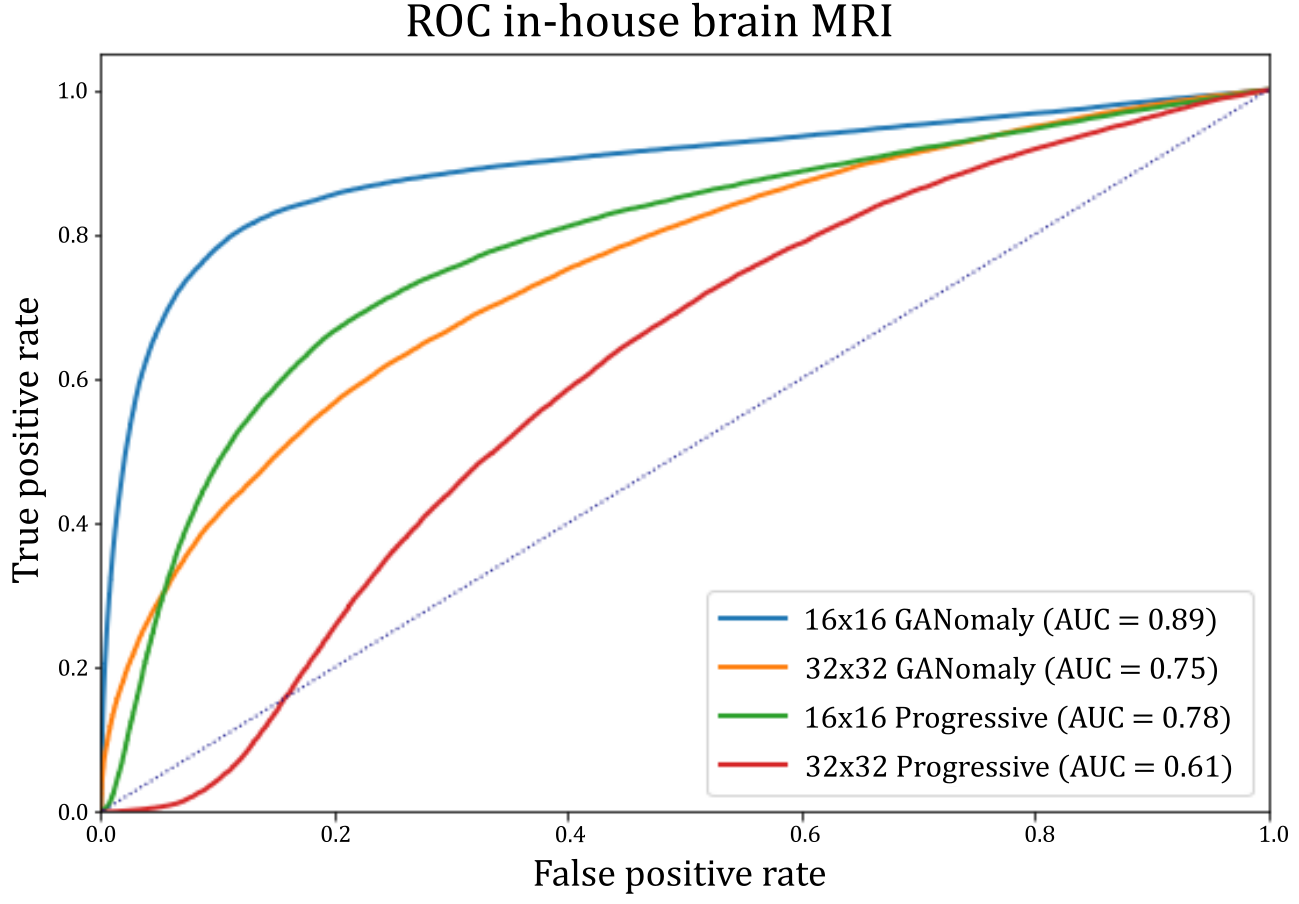}
	\caption{Receiver operating characteristic (ROC) curves for GANomaly and progressive GANomaly, using different patch sizes on the in-house brain MRI dataset with white matter hyperintensities. The x-axis shows the true positive rate and the y-axis shows the false positive rate.}
	\label{fig:rocmri}
\end{figure}

GANomaly with 16x16 patches obtains an AUC of 0.89. Progressive GANomaly with the same patches is less sensitive and less specific with an AUC of 0.78. GANomaly using the 32x32 patches performs comparably to 16x16 progressive GANomaly with an AUC of 0.75. Progressive GANomaly using 32x32 patches receives a score of 0.61 and does not seem to detect the anomalies. Figure \ref{fig:rocmri} shows the full ROC curves for both methods and patch sizes.

\section{Discussion}
In this work, a novel anomaly detection method is proposed by combining GANomaly with progressively growing GANs. The first experiment using Fashion MNIST has shown a better performance of GANomaly and progressive GANomaly compared to the baseline method, a one-class SVM. The methods reported an average F1-score of 91.8, 92.8 and 79.0, respectively. The second experiment used preprocessed brain images and compared the two GAN based methods with each other using different patch sizes. Two controlled tests have been performed using this dataset by creating toy images. The intensity test has shown that progressive GANomaly using 32x32 patches is the best all-round performer. During the size test 16x16 progressive GANomaly is able to achieve the highest sensitivity without compromising on specificity. On our in-house brain MRI dataset, GANomaly using the smaller patch sizes performed the best.

The results from the experiments performed on the Fashion MNIST dataset adhered to the hypothesis that items that are closer to each other in latent space are more difficult to distinguish. This is seen with the lower F1-scores for trying to find sneakers when trained on boots. During training, a difference has been observed between the final losses of GANomaly and progressive GANomaly. The latter is lower than the converged loss from the WGAN used in GANomaly after training. This means that the reconstructions and encodings are better on the train-validation set. The claim that the reconstructions are better using progressive GANomaly is confirmed by the higher SSIM scores of the MOOD patches compared to the SSIM scores of GANomaly. This is likely the cause for the higher performance using progressively growing GANs.

Different patch sizes were investigated using the MOOD dataset and the larger patch size seems to be more sensitive to intensity differences when a large anomaly is present. Progressive GANomaly performs better than GANomaly using the larger patch sizes, which is likely because of the greater capabilities of the progressively growing GAN during training. The smaller 16x16 patches performed better when trying to find anomalies that are smaller than 6~mm. Here, progressive GANomaly reached a slightly higher sensitivity compared to GANomaly.

In this work, the algorithms were tested only on the detection of WMHs during the test using our in-house brain MR images. However, anomaly detection is supposed to be a general detection system. Since the algorithms are most sensitive to bright lesions\cite{Meissen2021}, as seen in Table \ref{tab:mood}, the WMHs were chosen as a target. It would be interesting to further explore how this algorithm performs on different datasets and pathologies. According to the intensity test performed on the MOOD dataset, the proposed algorithm could be used in other brain pathologies such as tumors. 

This method is feasible for the use of anomaly detection, especially in case of large-throughput screening. (Progressive) GANomaly is a reasonable pre-analyzing method to reduce the workload of radiologists. Since this is an anomaly detection system meant to help radiologists, a high sensitivity is preferred over specificity. This would in practice be a method that highlights designated spots that need to be checked and verified by radiologists. However, progressive GANomaly does need some further research to prove that it performs better than GANomaly in other cases with different pathologies. Since it performs better using toy images, does not directly translate to the performance on real deformities, like seen in the WMH experiment on our in-house brain MRI. Certain parameters can also be further finetuned. In these experiments, the parameters that worked best on GANomaly were chosen to equalize the playing field. Finetuning these parameters to progressive GANomaly specifically might further improve the algorithm. 

Van Hespen et al. detected 97.5\% of total brain infarct volume with GANomaly\cite{Hespen2021}. That network has been trained on 697 subjects and was thus trained on more data compared to the in-house brain MRI experiment, which was trained on 90 subjects. A large number of patches was extracted from a limited number of subjects to provide enough data, resulting in overlapping patches. Progressive GANomaly trained with more data could increase the number of brain abnormalities detected. Especially since the lesions that were missed were smaller than 1 ml in volume. In the experiment on our in-house brain MRI data, it has also been observed that GANomaly misses the smallest WMHs. The MOOD size test shows that progressive GANomaly performs slightly better in case of smaller anomalies. Post-processing was also performed to optimize GANomaly as a detection method in the work of Van Hespen et al. This has not been applied to our experiments, however, it would be insightful to see how both methods would perform if the anomaly scores were to be processed to their fullest potential.

\subsubsection{Limitations}
The speed of GANomaly during training and computation of anomaly scores is about a ten-fold faster than progressive GANomaly. However, there is still room to optimize progressive GANomaly and the speed should not be a large limitation, since training of algorithms can be performed off-line.

It can be argued that the proposed progressive GANomaly is too sensitive to normal brain structures. This is observed when using the smaller patches: the method is inherently more sensitive to white matter, which is visualized in Figure \ref{fig:mood}. The image obtains a relatively higher anomaly score for the white matter compared to the bigger patch sizes. It is also seen in the results of the MOOD intensity test in Figure \ref{fig:roc}(a), where 16x16 progressive GANomaly performed the worst. It has been observed that the reconstructions can be a bottleneck in this method. If the model is able to reconstruct the anomaly perfectly, the encoder loss will be low. This results in the anomaly getting an unintentional low score. Using larger patches, the training finished with a higher contextual loss compared to the smaller patches. This translates into the quality of the reconstructions. This might be one of the underlying reasons for the larger patches to perform better in the MOOD intensity test. The same sensitivity is not observed when using larger patches. The advantage of using large patches is that more information on the structures can be taken into account. Nevertheless larger patches do have their own limitations on detecting anomalies that are smaller than 5~mm in diameter. 

Additionally, it must be noted that GANomaly was optimized during these tests, while progressive GANomaly could not utilize the same optimization approach. Both the median and median absolute deviation per element of the encoder loss become zero after training with progressively growing GANs. 

In the MOOD test, it has been observed that progressive GANomaly receives both the best and worst AUC scores. This shows that the patch sizes have a large effect on the capabilities of the algorithm. It would be optimal if the different patch sizes could work together to create an all round detector excelling at both varying intensities and sizes. 

GANomaly does seem to be more robust, since it performed better on our in-house brain MRI dataset. The heatmap in Figure \ref{fig:UDES}(c) shows the failure of progressive GANomaly in this experiment. This dataset is much smaller compared to the other tests and the `healthy' patches are not of the best quality, since the subjects are relatively old. Brains of elderly tend to have deformities that look similar to WMHs, for instance calcifications. Yet these are not labeled and end up in the ``normal'' training set. 

Progressive GANomaly needs to train for a longer time period and therefor needs more iterations through the dataset. Additionally, since progressive GANomaly consists out of multiple layers for each resolution, the network is composed of more convolutional layers and has more trainable parameters. This might be why the dataset size has a bigger effect on the proposed algorithm.

\section{Conclusion}
The substitution of the WGAN in GANomaly by progressively growing GANs enabled a lower loss at the end of training and is able to create more accurate reconstructions. The proposed method outperforms the current GANomaly architecture as an image-level anomaly detector, which has been demonstrated using Fashion MNIST; and additionally as a pixel-level anomaly detector using brain images on the condition that the training dataset is large enough. The experiments performed in this paper indicate that the use of progressively growing GANs have the potential to improve generalized adversarial training for anomaly detection.

\bibliography{report} 
\bibliographystyle{spiebib} 

\newpage
\appendixpage    
\begin{table*}[h]
\caption{Result of the experiment on Fashion MNIST. The boots are classified as the normal label and compared against dresses, sandals, and sneakers. The accuracy, sensitivity, precision, and F1-scores are reported for the methods one-class SVM, GANomaly, and progressive GANomaly.}
\begin{tabularx}{\textwidth}{@{}llYYYY@{}}
\toprule
\multicolumn{2}{c}{} & Accuracy & Sensitivity & Precision & F1-score\\
\midrule
 \multirow{3}{*}{Boots vs dress} & SVM & 74.6 & \textbf{100.0} & 66.3 & 79.8\\
& GANomaly & 99.5 & 99.2 & \textbf{99.7} & 99.5 \\
& Progressive & \textbf{99.6} & 99.7 & 99.6 & \textbf{99.6} \\
\midrule
\multirow{3}{*}{Boots vs sandals} & SVM & 74.5 & \textbf{99.8} & 66.3 & 79.7\\
& GANomaly & 94.8 & 94.9 & 94.7 & 94.8\\
& Progressive & \textbf{95.8} & 95.8 & \textbf{95.8} & \textbf{95.8}\\
\midrule
\multirow{3}{*}{Boots vs sneakers} & SVM & 72.5 & \textbf{95.8} & 65.4 & 77.7\\
& GANomaly & 81.5 & 80.1 & 82.4 & 81.2\\
& Progressive & \textbf{83.6} & 81.2 & \textbf{85.4} & \textbf{83.2}\\
\bottomrule
\end{tabularx}
\label{tab:fashion}
\end{table*}

 \begin{table*}[h]
\caption{AUC scores obtained from the intensity test and the size test. The two methods, GANomaly and progressive GANomaly, are compared. Each method is compared using two different patch sizes, 16x16 and 32x32. Left: the AUC score is given for varying intensities, with a disk of fixed size 35 mm. Right: the AUC score is given for varying disk sizes, with a fixed intensity of 1.0.}
\centering
\begin{tabularx}{\textwidth}{@{}rYYYYrYYYY@{}}
\toprule
\multicolumn{5}{c}{\bfseries Intensity test}&\multicolumn{5}{c}{\bfseries Size test}\\
\cmidrule(lr){1-5} \cmidrule(lr){6-10}
&\multicolumn{2}{c}{\bfseries GANomaly}
&\multicolumn{2}{c}{\bfseries Progressive} &&\multicolumn{2}{c}{\bfseries GANomaly}
&\multicolumn{2}{c}{\bfseries Progressive} \\
\cmidrule(lr){2-3} \cmidrule(l){4-5} \cmidrule(lr){7-8} \cmidrule(l){9-10} 
\bfseries Intensity
&\bfseries 16x16 
&\bfseries 32x32 
&\bfseries 16x16  
&\bfseries 32x32
& \bfseries Size (mm)
&\bfseries 16x16 
&\bfseries 32x32 
&\bfseries 16x16  
&\bfseries 32x32\\
\midrule
\bfseries 0.1 & 0.61 & 0.79 & 0.06 & 0.57 & \bfseries 1 &0.37&0.37&0.91&0.21\\
\bfseries 0.2 & 0.39 & 0.38 & 0.17 & 0.79 & \bfseries 2 &0.77&0.47&0.87&0.25\\
\bfseries 0.3 & 0.10 & 0.14 & 0.12 & 0.13 & \bfseries 3 &0.93&0.65&0.96&0.62\\
\bfseries 0.4 & 0.07 & 0.08 & 0.12 & 0.56 & \bfseries 4 &0.99&0.91&0.99&0.74\\
\bfseries 0.5 & 0.15 & 0.31 & 0.99 & 0.88 & \bfseries 5 &1.00&0.99&1.00&0.88\\
\bfseries 0.6 & 0.60 & 0.88 & 0.62 & 0.97 & \bfseries 6 &1.00&1.00&1.00&0.95\\
\bfseries 0.7 & 0.94 & 1.00 & 0.39 & 0.97 & \bfseries 7 &1.00&1.00&1.00&0.99\\
\bfseries 0.8 & 1.00 & 1.00 & 0.99 & 1.00 & \bfseries 8 &1.00&1.00&1.00&1.00\\
\bfseries 0.9 & 1.00 & 1.00 & 1.00 & 1.00 & \bfseries 9 &1.00&1.00&1.00&1.00\\
\bfseries 1.0 & 1.00 & 1.00 & 1.00 & 1.00 & \bfseries 10 &1.00&1.00&1.00&1.00\\
\bottomrule
\end{tabularx}
\label{tab:mood}
\end{table*}

\end{document}